\pdfoutput=1

\documentclass[11pt,a4paper]{article}
\usepackage{acl}

\usepackage[most]{tcolorbox}
\tcbuselibrary{skins,breakable}
\usepackage{wrapfig}

\usepackage{times}
\usepackage{latexsym}
\usepackage{graphicx}
\usepackage{booktabs}
\usepackage[T1]{fontenc}

\usepackage[utf8]{inputenc}

\usepackage{microtype}
\usepackage{inconsolata}

\usepackage{graphicx}
\usepackage{fancyhdr}
\usepackage{dblfloatfix} 
\usepackage{float}
\usepackage{pifont}      
\usepackage{array}       
\usepackage{makecell}    
\usepackage{ragged2e}    
\usepackage{rotating}
\usepackage{tabularx}
\usepackage{colortbl}
\usepackage{multirow}
\usepackage{enumerate}
\usepackage{longtable}
\usepackage{url}         
\usepackage{pdflscape}   
\usepackage{float}       
\usepackage{graphicx}
\usepackage{subcaption}

\newcolumntype{P}[1]{>{\centering\arraybackslash}p{#1}}
\newcolumntype{M}[1]{>{\centering\arraybackslash}m{#1}}
\newcolumntype{L}[1]{>{\raggedright\arraybackslash}m{#1}}
\newcolumntype{C}[1]{>{\centering\arraybackslash}m{#1}}

\usepackage{xcolor}
\usepackage{enumitem}
\setlist[itemize]{leftmargin=0.2em}

\setlist[description]{leftmargin=1.2em,labelsep=0.4em,itemsep=2pt,topsep=2pt,font=\bfseries}

\usepackage{xcolor}
\definecolor{darkgreen}{rgb}{0.0,0.4,0.0}

\title{APEX\textendash Agents}

\author{
\textbf{Bertie Vidgen} \quad \textbf{Austin Mann} \quad \textbf{Abby Fennelly} \quad \textbf{John Wright Stanly} \\
\textbf{Lucas Rothman} \quad \textbf{Marco Burstein} \quad \textbf{Julien Benchek} \quad \textbf{David Ostrofsky} \\
\textbf{Anirudh Ravichandran} \quad \textbf{Debnil Sur} \quad \textbf{Neel Venugopal} \quad \textbf{Alannah Hsia} \\
\textbf{Isaac Robinson} \quad \textbf{Calix Huang} \quad \textbf{Olivia Varones} \quad \textbf{Daniyal Khan} \quad \\
\textbf{Michael Haines} \quad \textbf{Austin Bridges} \quad \textbf{Jesse Boyle} \quad \textbf{Koby Twist} \\
\textbf{Zach Richards} \quad \textbf{Chirag Mahapatra} \quad \textbf{Brendan Foody} \quad \textbf{Osvald Nitski} \\
\vspace{0.5em}
Mercor \quad apex@mercor.com
}

\begin{document}
\pagestyle{fancy}
\fancyhf{}  

\fancyhead[L]{\includegraphics[width=\headwidth]{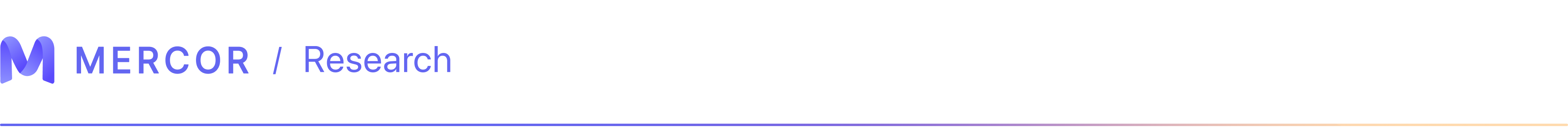}}

\fancyfoot[C]{\thepage}

\renewcommand{\headrulewidth}{0pt}

\fancypagestyle{plain}{
  \fancyhf{}
  \fancyhead[L]{\includegraphics[width=\headwidth]{images/banner.png}}
  \fancyfoot[C]{\thepage}
  \renewcommand{\headrulewidth}{0pt}
}

\maketitle
\thispagestyle{fancy}

\setlength{\parindent}{0pt}

\begin{abstract}
We introduce the AI Productivity Index for Agents (\textbf{APEX\textendash Agents}), a benchmark for assessing whether AI agents can execute long-horizon, cross-application tasks created by investment banking analysts, management consultants, and corporate lawyers. APEX\textendash Agents requires agents to navigate realistic work environments with files and tools. We test eight agents for the leaderboard using Pass@1. Gemini 3 Flash (Thinking=High) achieves the highest score of $24.0\%$, followed by GPT-5.2 (Thinking=High), Claude Opus 4.5 (Thinking=High), and Gemini 3 Pro (Thinking=High). We open source the APEX\textendash Agents benchmark ($n=480$) with all prompts, rubrics, gold outputs, files, and metadata. We also open source \textbf{Archipelago}, our infrastructure for agent execution and evaluation.
\end{abstract}

\begin{figure}[t]
\centering
\includegraphics[width=1\linewidth]{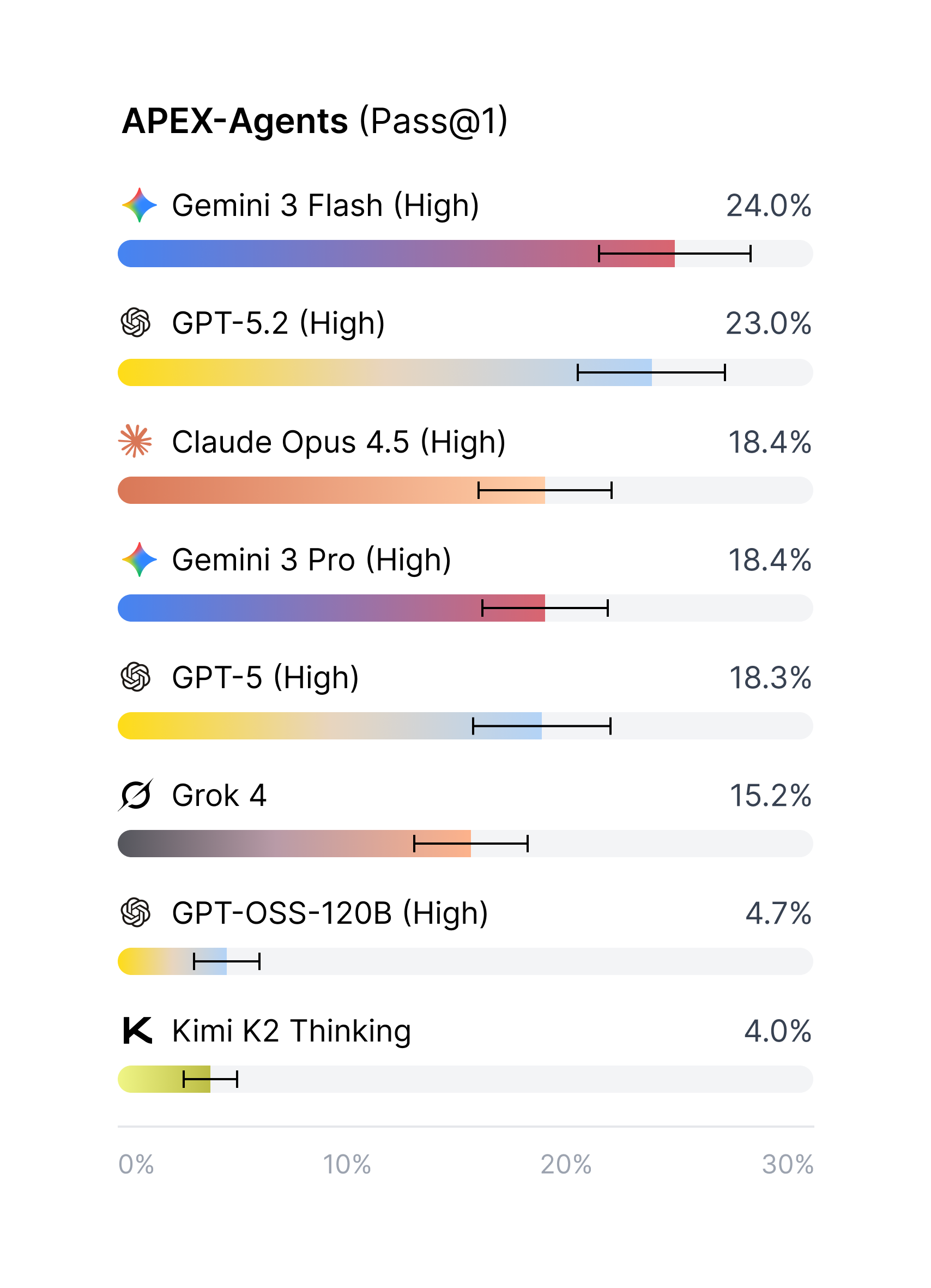}
\caption{Performance of agents on the APEX\textendash Agents benchmark using Pass@1. Thinking settings are in parentheses.}
\label{fig:benchmark_overall_performance}
\end{figure}

\begin{figure*}[t]
\centering
\includegraphics[width=1\linewidth]{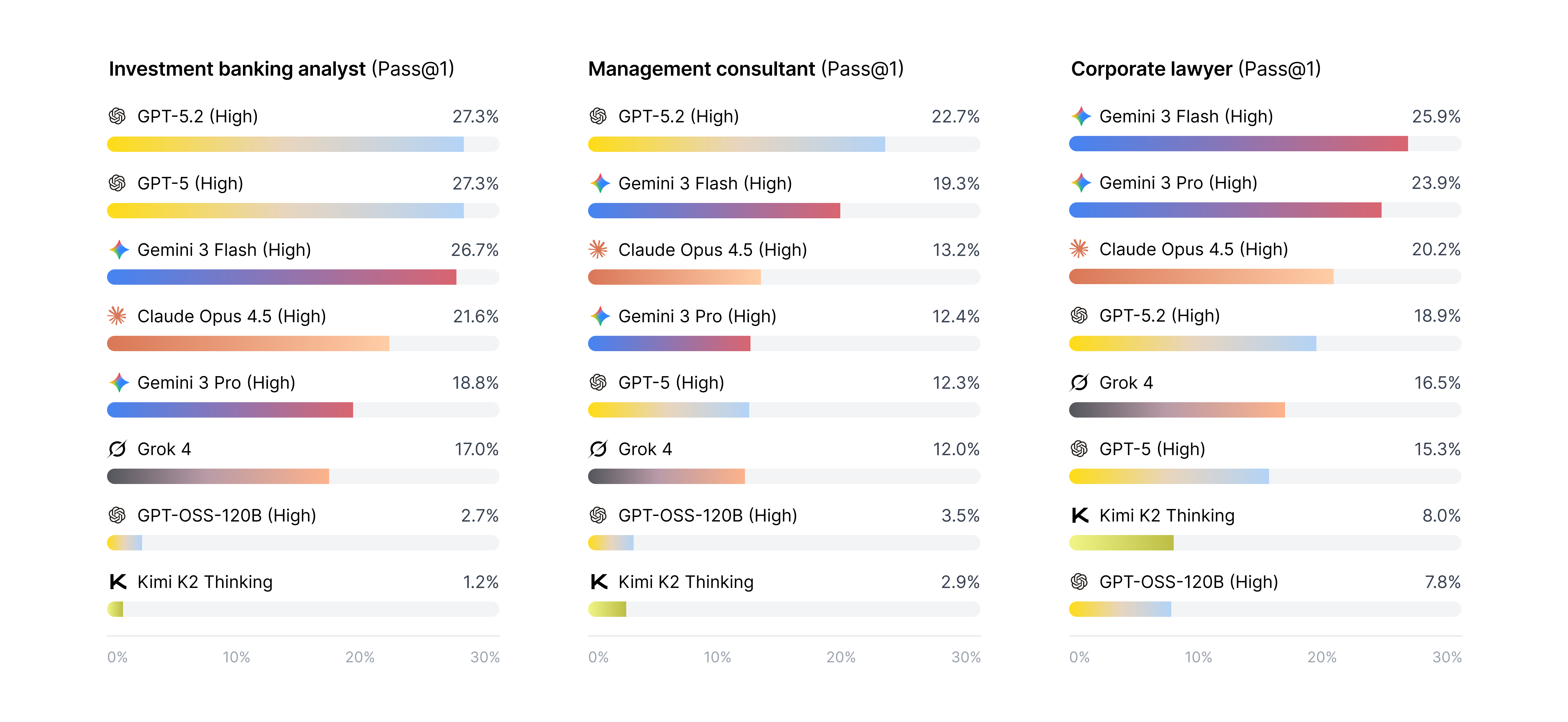}
\caption{Performance of agents on the APEX\textendash Agents benchmark using Pass@1, segmented by job. Thinking settings are in parentheses.}
\label{fig:final_segmented_leaderboard}
\end{figure*}

\section{Introduction}
If AI agents can reliably execute professional services work, the economic and social consequences will be profound. A large team of skilled, diligent experts will be available on-demand to anyone, fundamentally reshaping knowledge work and dramatically increasing productivity. Enterprises such as Box, Salesforce, and Databricks are beginning to deploy agentic systems at scale, and AI labs are investing substantial resources in expanding agentic capabilities.
\vspace{1em}

Agents present a new frontier for AI -- yet existing agentic evals have a large sim-to-real gap, and do not capture how professionals work day-to-day~\citep{kapoor2024aiagentsmatter, styles2024workbench, dechezelles2025browsergymecosystemwebagent, froger2025arescalingagentenvironments, meimandi2025measurementimbalanceagenticai, vidgen2025aiproductivityindexapex}. They are often narrowly scoped, highly contrived, and contain only simple tasks, providing limited signals into how agents can help professionals in the real world.
\vspace{1em}

To assess whether AI agents can execute highly complex professional services work, we present APEX\textendash Agents, a new benchmark for frontier AI evaluation. The tasks were created by investment banking analysts, management consultants, and corporate lawyers, and require agents to reason, demonstrate advanced knowledge, use multiple applications, and plan over long horizons.
\vspace{1em} 

APEX\textendash Agents was built in three steps. First, we created data-rich worlds, each based on a unique project scenario. Industry professionals were assigned to teams, given roles (e.g., partner, associate), and tasked with delivering the project over 5-10 days. They planned work, conducted research, and produced high-quality customer-ready deliverables from scratch. Second, professionals created realistic and challenging tasks using the files from within each world. Third, we gave agents access to each world so they could execute the tasks (with all of the data and software that a human would use). This approach is inspired by findings from the APEX Survey (Section~\ref{sec:apex_survey}) and feedback from industry experts and our partners. Tasks are complex and, on average, experienced professionals estimated they take $1$–$2$ hours to complete.
\vspace{1em}

APEX\textendash Agents contains $480$ tasks, split across $33$ worlds.  We tested eight models (see Figure~\ref{fig:benchmark_overall_performance}). Gemini 3 Flash performs best, scoring $24.0\%$ on Pass@1, followed by GPT-5.2 at $23.0\%$. The difference between these models is not statistically significant. Claude Opus 4.5 and Gemini 3 Pro follow at $18.4\%$ and $18.4\%$, respectively.\footnote{Where available, we implement models with thinking / reasoning effort set to high. In the remainder of the paper we refer to models by their names alone.} The two open-source models we tested performed much worse than the closed-source models, scoring under $5\%$. Tasks are split evenly across the three jobs ($n=160$ each). They vary in difficulty, with the top-performing model scoring $27.3\%$ for investment banking tasks (GPT-5 and GPT-5.2), $22.7\%$ for management consulting (GPT-5.2) and $25.9\%$ for corporate law (Gemini 3 Flash). See Figure~\ref{fig:final_segmented_leaderboard}.
\vspace{1em}

The APEX\textendash Agents dataset is on Hugging Face with a CC-BY license.\footnote{\href{https://huggingface.co/datasets/mercor/apex-agents}{huggingface.co/datasets/mercor/apex-agents}} Archipelago, our infrastructure for agent execution and evaluation, is also open-source.\footnote{\href{https://github.com/Mercor-Intelligence/archipelago}{github.com/Mercor-Intelligence/archipelago}}

\begin{figure*}[t]
\centering
\includegraphics[width=1\linewidth]{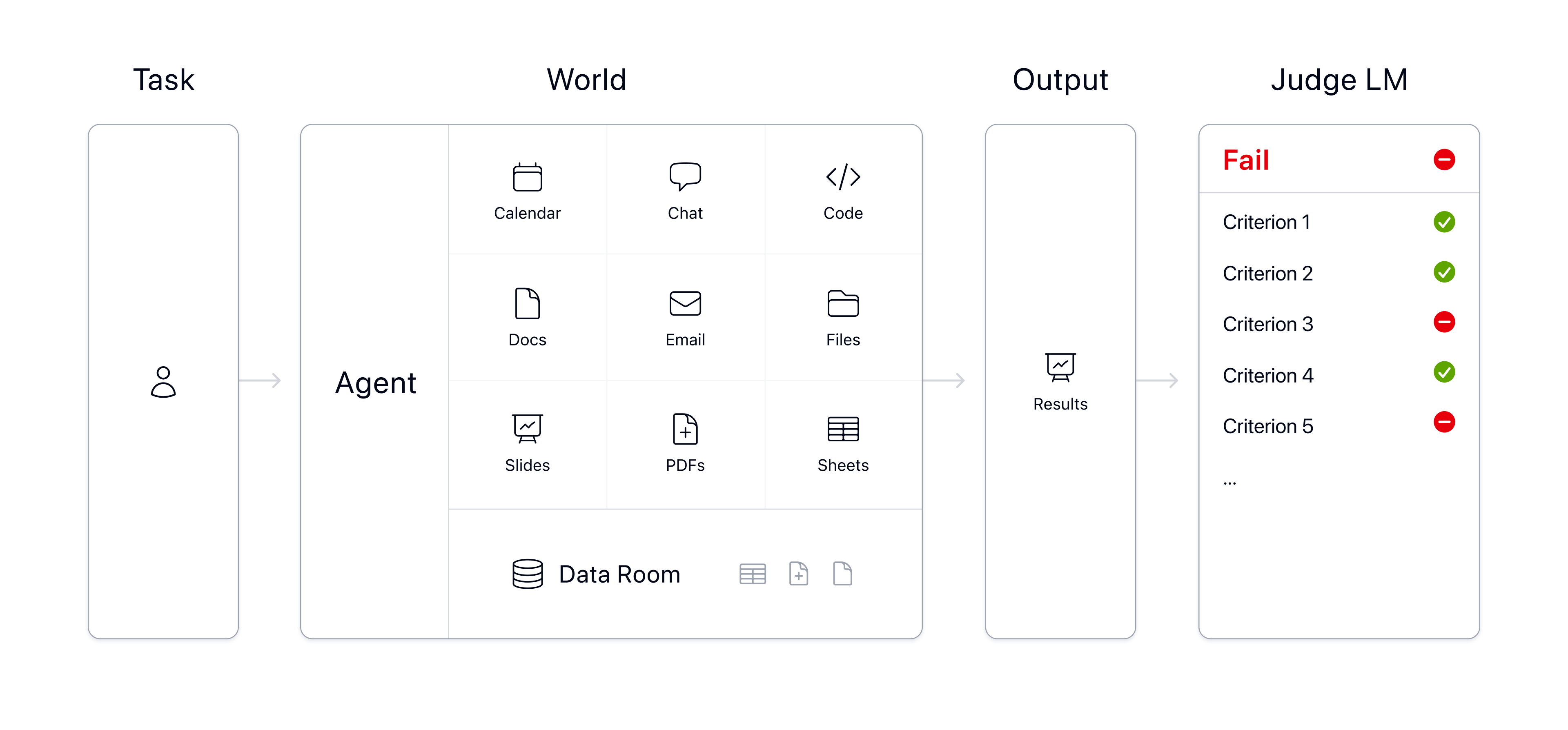}
\caption{Overview of the approach in APEX\textendash Agents to agentic AI evaluation.}
\label{fig:apex_process_diag}
\end{figure*}

\section{APEX Survey} \label{sec:apex_survey}
We surveyed $227$ experts from the Mercor talent marketplace to understand their day-to-day work, informing the creation of worlds and tasks in APEX\textendash Agents. Participants are a diverse representation of elite professionals, including $58$ financial and investment analysts (O*NET 13-2051), $77$ management consultants (O*NET 13-1111), and $92$ lawyers (O*NET 23-1011). On average, they had $10.8$ years of professional experience.

\paragraph{Work activities}
We asked participants what percentage of their time is spent on core activities, learning, admin, communications, meetings, and non-productive time. Results for each job are shown in Table~\ref{tab:time_spent} in the Appendix, with core activities comprising $47\%$ of the total. We asked participants to segment how they spend their time on core activities, providing a free text description of each activity and a percentage time allocation. We manually reviewed these descriptions and grouped them into one of 18 inductively-identified categories. This analysis was shared with contributors, and informed task creation for the APEX\textendash Agents benchmark. The five most frequent categories for each job are shown in Table~\ref{tab:time_spent_detailed} in the Appendix.
\vspace{1em}

\section{Benchmark design and dataset}
The composition of the APEX\textendash Agents benchmark is described in Table~\ref{tab:dataset_overview} and an overview of our approach is shown in Figure~\ref{fig:apex_process_diag}.
\vspace{1em}

\subsection{World creation}
Each world starts from a project scenario created by experts, and inspired by projects from their professional work. The scenario provides context about the delivery company, the customer, the project background and goals, and relevant constraints, such as:

\begin{quote}
A team of consultants from NorthPoint Strategy Partners are working on behalf of PureLife Wellness to develop a five-year market expansion and growth strategy by mapping global demand, assessing consumer trends, evaluating market entry options, and prioritizing high-potential international markets and product opportunities.
\end{quote}
\vspace{1em}

After each scenario was finalized,  were assigned in-world roles (e.g., partner, analyst, associate). They assumed role-appropriate responsibilities, working toward the project objectives with similar information, challenges, and trade-offs as an actual customer engagement. They sent emails and messages, conducted research, scoped deliverables, and iterated on the work products such as spreadsheets, insight reports, and slide decks. For each project, one expert played the role of the customer.
\vspace{1em}

The benchmark comprises $33$ worlds: $10$ in investment banking, $11$ in management consulting, and $12$ in law. On average, each world contains $166$ files. Of the $33$ worlds, $22$ involve entirely fictional entities, $9$ involve real companies placed in fictional scenarios, and $2$ contain both real and fictional companies. Each world  has nine applications, comprising $63$ tools: Calendar, Chat, Code Execution, Documents, File system, Mail, PDFs, Spreadsheets, and Presentations. Two investment banking worlds also have EDGAR SEC; an application for data on Equities and Financial Markets; and an application for data on Fixed Income Markets, adding $187$ tools. Web search is turned off for the benchmark to keep evaluations reproducible. To maintain validity, worlds contain all files required to execute tasks. 

\vspace{1em}

\begin{table*}[t]
\small
\centering
\caption{Overview of the APEX\textendash Agents benchmark dataset. The average estimated hours to complete tasks are provided by the experts. See Section~\ref{sec:baselining} for the results of the Baselining study.}
\label{tab:dataset_overview}
\begin{tabular}{l|cccccc}

\toprule
\textbf{Split} &
\makecell{\textbf{Number of} \\  \textbf{worlds}} &
\makecell{\textbf{Avg files} \\ \textbf{per world}} &
\makecell{\textbf{Number of} \\  \textbf{tasks}} &
\makecell{\textbf{Avg criteria} \\  \textbf{per task}} &
\makecell{\textbf{Avg estimated} \\  \textbf{hours to complete}} &
\makecell{\textbf{Tasks with}   \\  \textbf{file outputs}} \\
\midrule
Investment banking      & $10$     & $172$  & $160$  & $2.93$  & $1.36$  & $27$ ($16.9\%$) \\
Law                     & $12$     & $161$  & $160$  & $4.57$  & $2.40$  & $20$ ($12.5\%$) \\
Management consulting   & $11$     & $165$  & $160$  & $4.68$  & $1.69$  & $11$ ($6.9\%$)  \\
\midrule
Benchmark               & $33$     & $166$  & $480$  & $4.06$   & $1.82$ & $58$ ($12.1\%$) \\
\bottomrule
\end{tabular}
\end{table*}

\subsection{Tasks}
Once each world was completed, experts created long-horizon tasks that are realistic, challenging, and diverse. Worlds have between $8$ and $20$ tasks, with $14.5$ on average. Each task is provided as a single-turn prompt and can only be executed by using the in-world files and applications. Tasks were informed by the APEX Survey (Section~\ref{sec:apex_survey}). When creating tasks, contributors selected from a pool of frontier agents (GPT-5, Claude Opus 4.5, Gemini 3 Pro) to collect outputs, allowing them to adversarially iterate. Tasks underwent multiple rounds of review during production.
\vspace{1em}

The task prompts make explicit what output is required from the agent. $422/480$ tasks require a message in the console. For the other $58$, the agent has to either create spreadsheets ($14$), documents ($20$) or presentations ($5$), or edit existing spreadsheets ($16$), documents ($2$) or presentations ($1$).  An example task, selected for brevity, is given below.

\begin{quote}
Reply back to me with the P/E ratio for KVUE, rounded to two decimal points. Use the implied share price in the DCF model and diluted EPS from the annual financials dated 12/23/2025.
\end{quote}
\vspace{1em}

\subsection{Rubrics and gold outputs}
Experts created rubrics to grade agents' output. Rubrics contain criteria: self-contained, short, descriptive statements, which can be graded as true or false against an agent's output. Each criterion also has a ``grading target'', which specifies the type of output required by the prompt. Criteria only reward critical aspects of the output, i.e., elements that are required for the task to be considered complete.
\vspace{1em}

There are between $1$ and $10$ criteria per task, with a mean of $4.06$. Investment banking has fewer criteria than law and management consulting, on average, because there are more tasks that require just a single value to be returned (e.g., an updated EBITDA). These single value outputs can only be achieved by meeting many constraints and updating numerous intermediary values. 
\vspace{1em}

For each task, experts also created gold outputs. Gold outputs contain the exact information to address the prompt and are in the output type requested (e.g., message, document). Experts manually graded gold outputs against the rubric when creating them, ensuring prompt--rubric alignment. Experts also tagged each prompt and rubric with metadata, including the expected output type, workflow, and expected time for an industry professional to complete the task in the real-world. The number of tasks associated with each workflow is in Table~\ref{tab:workflow_distribution} in the Appendix. 
\vspace{1em}

\subsection{Contributors to APEX\textendash Agents}
$256$ experts contributed to APEX\textendash Agents, sourced from Mercor's talent marketplace. Experts' mean experience is $12.9$ years and the median is $11.0$ years. Experts include former consultants from BCG and McKinsey, investment bankers from Morgan Stanley and Citigroup, and corporate lawyers at Disney and other Fortune 500 companies. Experts worked as both contributors and reviewers; creating project scenarios, building worlds, creating tasks, auditing quality, and checking gold outputs and rubrics.
\vspace{1em}

\subsection{Baselining study}\label{sec:baselining}
For $20\%$ of the tasks ($n=96$), experts who had not created or reviewed them independently executed them from scratch. This checks (1) whether the tasks can actually be completed, (2) the fairness of the rubric, and (3) the time estimates provided by the experts. In $10\%$ of tasks, we identified a minor problem with the prompt, rubric, or metadata that needed to be fixed. We then cascaded these fixes to the rest of the dataset. For these sample tasks, experts estimated the time to complete the tasks at $1.70$ hours (note that the estimate for the whole benchmark is $1.82$ hours) whereas the true time was $1.37$ hours. This is an over-estimate of $0.45$ hours or $33\%$.
\vspace{1em}

\begin{table}[b]
\small
\centering
\caption{Confusion matrix for the judge model, evaluated against human-labeled ground truth ($n=747$ labels, based on $n=249$ criteria).}
\label{tab:rubric_ground_truth}
\begin{tabular}{lcc}
\toprule
 & \multicolumn{2}{c}{\textbf{Predicted}} \\
\cmidrule(lr){2-3}
\textbf{Actual}   &  \textbf{Met}   &   \textbf{Failed}  \\
\midrule
\textbf{Met}      &  \textbf{208}   &   4                \\
\textbf{Failed}   &  7              &   \textbf{528}     \\
\bottomrule
\end{tabular}
\end{table}

\begin{table*}[b]
\small
\centering
\caption{Performance of agents on the APEX\textendash Agents benchmark. Where available, models have thinking / reasoning effort set to high.}
\label{tab:benchmark_result_overall}
\begin{tabular}{l|c|ccc|ccc}
\toprule
\textbf{Model} &
\textbf{Pass@1} &
\textbf{Pass@8} &
\textbf{Pass\textasciicircum8} &
\makecell{\textbf{Mean} \\  \textbf{score}} &
\makecell{\textbf{IB analyst} \\  \textbf{Pass@1}} &
\makecell{\textbf{Consultant} \\  \textbf{Pass@1}} &
\makecell{\textbf{Lawyer} \\  \textbf{Pass@1}} \\
\midrule
Claude Opus 4.5    & $18.4\%$ [$15.5$--$21.3$]           & $34.0\%$ [$29.8$--$38.3$]                 & $8.8\%$             & $34.8\%$           & $21.6\%$           & $13.2\%$             & $20.2\%$ \\
Gemini 3 Flash     & \textbf{$24.0\%$ [$20.7$--$27.3$]}  & $36.7\%$ [$32.3$--$41.0$]                 & \textbf{$13.4\%$}   & \textbf{$39.5\%$}  & $26.7\%$           & $19.3\%$             & \textbf{$25.9\%$} \\
Gemini 3 Pro       & $18.4\%$ [$15.7$--$21.1$]           & $37.3\%$ [$32.9$--$41.7$]                 & $6.5\%$             & $34.1\%$           & $18.8\%$           & $12.4\%$             & $23.9\%$ \\
GPT-5              & $18.3\%$ [$15.4$--$21.3$]           & $31.0\%$ [$26.9$--$35.4$]                 & $7.7\%$             & $32.9\%$           & \textbf{$27.3\%$}  & $12.3\%$             & $15.3\%$ \\
GPT-5.2            & $23.0\%$ [$19.8$--$26.2$]           & \textbf{$40.0\%$ [$35.6$--$44.4$]}        & $11.0\%$            & $38.7\%$           & \textbf{$27.3\%$}  & \textbf{$22.7\%$}    & $18.9\%$ \\
GPT-OSS-120B       & $4.7\% $ [$3.3$--$6.1$]             & $11.5\%$ [$8.8$--$14.4$]                  & $1.2\%$             & $14.5\%$           & $2.7\%$            & $3.5\%$              & $7.8\%$\\
Grok 4             & $15.2\%$ [$12.8$--$17.7$]           & $32.9\%$ [$28.7$--$37.3$]                 & $4.7\%$             & $30.3\%$           & $17.0\%$           & $12.0\%$             & $16.5\%$ \\
Kimi K2 Thinking   & $4.0\% $ [$2.9$--$5.2$]             & $14.4\%$ [$11.5$--$17.5$]                 & $0.3\%$             & $11.5\%$           & $1.2\%$            & $2.9\%$              & $8.0\%$\\
\bottomrule
\end{tabular}
\end{table*}

\section{Evaluation}

\subsection{Collecting agent outputs}
Each agent executes each task eight times, yielding a total of $30,720$ trajectories ($8$ agents × $8$ runs × $480$ tasks). Trajectories comprise multiple steps taken by an agent, where each step involves using a tool or reflecting / planning. We apply a maximum of 250 steps per task, marking any trajectories that exceed it as a failure. Inspection of trajectories showed that, beyond this, the agent is typically stuck in a loop.
\vspace{1em}

Outputs are collected from closed-source models via their respective APIs. For open-source models, we use Baseten. LiteLLM is used as a wrapper to handle calls uniformly. See Appendix~\ref{appendix:model_configs} for information on model configs. We lightly optimized the system prompt used by the agents to help them execute tasks. It is given in Appendix~\ref{appendix:agent_setup}, along with details on the ReAct toolbelt used by the agents. 
\vspace{1em}

\subsection{Judge model}
Agents’ outputs are graded against the task rubric, with each criterion graded independently by a judge model. This results in $\sim\!\!125{,}000$ grades ($30,720$ trajectories x $4.06$ criteria). Based on several rounds of internal testing, we use Gemini 3 Flash, with thinking set to low, as the judge. The judge takes in the task prompt, the agent output, a log of the changes induced by the agent's actions, and the criterion -- but not the agent trajectory. It returns a binary score per criterion (Met, Not met) and a concise free text explanation. To identify what needs grading, such as a message printed to the console or an edited spreadsheet, we use an auxiliary judge that identifies the right artifact from the ``grading target'' associated with each criterion. The content of files is extracted using Reducto.\footnote{\href{https://reducto.ai/}{reducto.ai}}
\vspace{1em}

To assess the performance of the judge model, we constructed a ground truth eval set of labels for $60$ tasks ($20$ per job), comprising $249$ criteria (with each task corresponding to, on average, 4.15 criteria). For each task, we sampled final outputs from three of the models, resulting in $747$ criteria ($3$ x $249$). The ground truth labels were created independently, with experts not able to see LM judge grades. On a per-criterion basis, the eval set contains $212$ passes ($28.4\%$) and $535$ fails ($71.6\%$), similar to the distribution of criterion scores in the benchmark. 
\vspace{1em}

Judge model grades are shown in Table~\ref{tab:rubric_ground_truth}. Accuracy is $98.5\%$, precision is $96.7\%$, recall is $98.1\%$, and Positive-class F1 is $97.4\%$. This corresponds to a false negative rate of $1.9\%$ and a false positive rate of $1.3\%$. Given this, differences in benchmark scores that are below 1 percentage point should be interpreted cautiously. After the benchmark was finalized, we ran the judge model over the $n=480$ gold outputs and it graded all of the associated criteria correctly.
\vspace{1em}

Gemini 3 Flash serves as the judge model and is also evaluated on the leaderboard, creating a risk of self-preference. This is mitigated by the judge not viewing the trajectories when grading. Self-preference can be lower when grading well-defined criteria in a rubric, rather than abstract concepts like ``helpfulness'' or ``preference''. Out of $84$ criteria for Gemini 3 Flash in the ground truth eval set, the judge returned $1$ false positive. This equates to a false positive rate of $1.2\%$. Acknowledging the small sample, this rate is in line with the other agents we tested.
\vspace{1em}

\subsection{Metrics}
The APEX\textendash Agents leaderboard uses Pass@1, the proportion of tasks where the agents' outputs meet all criteria. We use Pass@1 because the criteria are all must-haves -- if any are not met, the task is incomplete. Pass@1 answers a clearly-defined question: if you select a task uniformly from the benchmark data and run an agent once, what is the probability that it passes (i.e., meets all criteria)?
\vspace{1em}

We collect 8 trajectories for each agent–task pair, and score each trajectory as pass or fail. For each agent and task, we compute the pass rate and report Pass@1 as the task-uniform mean of these per-task pass rates across the 480 tasks. We compute 95\% confidence intervals using task-level bootstrapping: we resample tasks with replacement ($n=480$) and recompute the task-uniform mean over $10,000$ resamples.
\vspace{1em}

Beyond the leaderboard, we report Pass@8 as an indicative ceiling on current frontier agent capabilities, showing whether the agent passes at least once when given eight attempts. We also report Pass\textasciicircum{$8$}, which assesses consistency by measuring whether the agent produces a correct output on every attempt out of eight. For $k<8$, we estimate Pass\textasciicircum{$k$} per task by sampling without replacement from each agent’s observed runs. Finally, we report the mean percentage of criteria passed per task. For training, this percentage can be more informative than Pass@1 because it provides a dense signal that rewards partial progress.
\vspace{1em}

\begin{figure*}[t]
\centering
\includegraphics[width=0.9\linewidth]{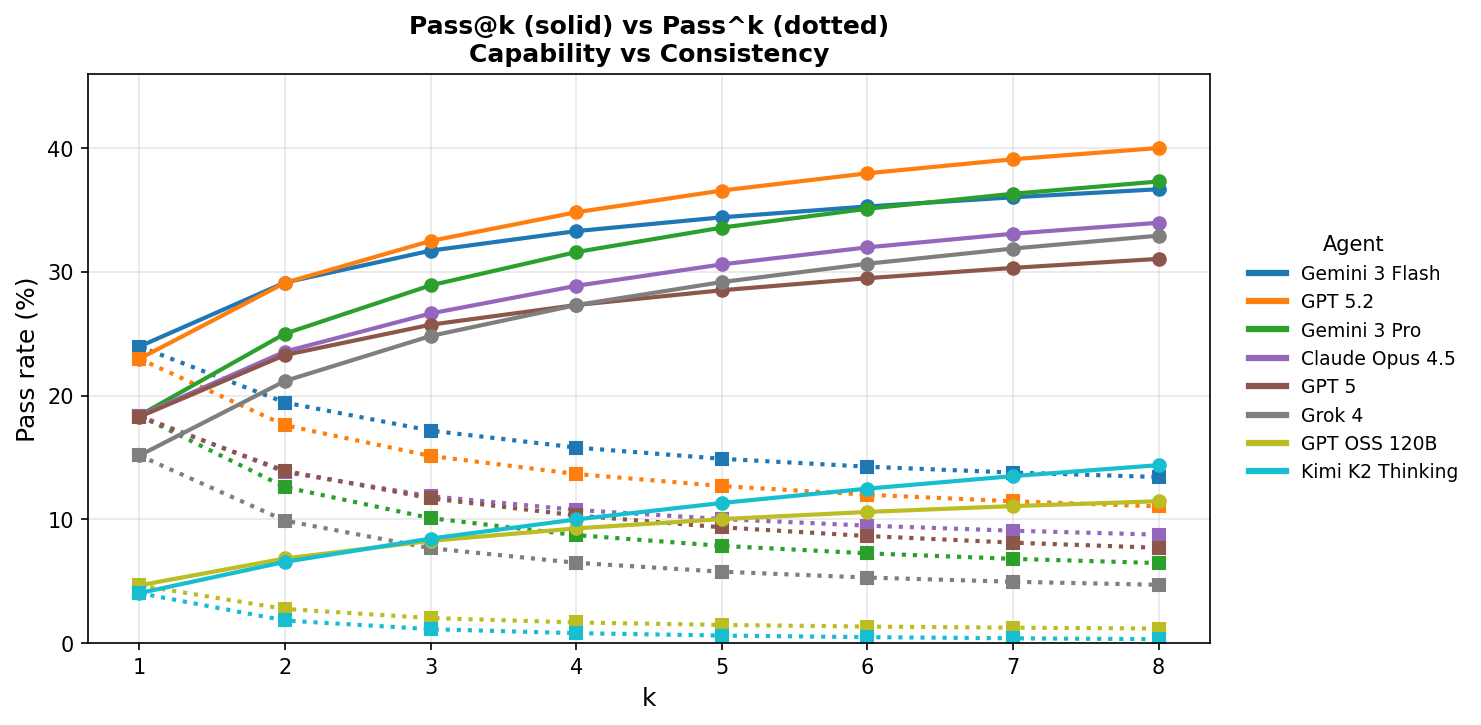}
\caption{Pass@k and Pass\textasciicircum{k} of agents on the APEX\textendash Agents benchmark. Solid lines indicate Pass@k, and dotted lines indicate Pass\textasciicircum{k}.}

\label{fig:k_analysis}
\end{figure*}

\section{Results on APEX\textendash Agents}

\paragraph{Pass@1 (Leaderboard)}
Agents' performance on Pass@1 is shown in Table~\ref{tab:benchmark_result_overall} and Figure~\ref{fig:benchmark_overall_performance}. Gemini 3 Flash performs best, scoring $24.0\%$. The next best agents are GPT-5.2, Claude Opus 4.5, and Gemini 3 Pro at $23.0\%$, $18.4\%$, and $18.4\%$ respectively. The open-source agents perform substantially worse than the closed-source ones, both scoring under $5\%$.
\vspace{1em}
We use McNemar’s exact tests to evaluate whether Pass@1 differences between agents are statistically significant. A Benjamini–Hochberg correction controls the false discovery rate at $5\%$. Under this correction, the difference between the two best-performing agents, Gemini 3 Flash and GPT-5.2, is not significant, but both models are significantly better than all other models. All pairwise differences between the commercial agents and the two open-source agents are significant, and the difference between the two open-source agents is not. See Section~\ref{sec:sig_tests} in the Appendix.
\vspace{1em}

\paragraph{Pass@1 by job}
The top-performing agent with Pass@1 for investment banking analyst tasks (GPT-5 and GPT-5.2, tied) scores $27.3\%$, for management consulting tasks scores $22.7\%$ (GPT-5.2), and for corporate lawyer tasks scores $25.9\%$ (Gemini 3 Flash). For all jobs, the two open-source agents score lowest. They are comparatively stronger in corporate law, scoring $7.8\%$ (GPT-OSS-120B) and $8.0\%$ (Kimi K2 Thinking).
\vspace{1em}

\paragraph{Pass@8}
The top-performing agent with Pass@8 is GPT-5.2 at $40.0\%$, followed by Gemini 3 Pro at $37.3\%$, and Gemini 3 Flash at $36.7\%$. This is approximately $15$ percentage points higher than Pass@1, and shows agents have capabilities but are inconsistent. We find limited evidence of saturation at 8 runs, with agents' scores increasing by no more than $1$ percentage point compared to 7. See Figure~\ref{fig:k_analysis}.
\vspace{1em}

\paragraph{Pass\textasciicircum{8}}
The top-performing agent with Pass\textasciicircum{$8$} scores $13.4\%$ (Gemini 3 Flash). Across commercial agents, Pass\textasciicircum{$8$} drops by approximately 10–12 percentage points relative to Pass\textasciicircum{1}. Agents' rank ordering remains unchanged as $k$ increases, suggesting there are limited differences in how consistent agents are.
\vspace{1em}

\paragraph{Mean score}
Gemini 3 Flash has the highest mean score at $39.5\%$, followed by GPT-5.2 ($38.7\%$) and Claude Opus 4.5 ($34.8\%$). Mean scores are similar to Pass@8 and much higher than Pass@1, showing that even when they fail to completely execute a task, agents can still create partially useful outputs. Across the eight runs from each agent on each task, the mean standard deviation is $0.14$, ranging from $0.09$ for GPT-OSS-120B to $0.16$ for Grok 4.
\vspace{1em}

\begin{table*}[b]
\small
\centering
\caption{The mean number of steps, tools, and tokens used by agents ($n=3,840$). Answer tokens are a subset of completion tokens. We round values over $1,000$ to the nearest hundred and over $10,000$ to the nearest thousand.}
\label{tab:trajectory_statistics}
\begin{tabular}{l|cc|cccc}
\toprule
\textbf{Model}
& \textbf{Steps}
& \textbf{Tools}
& \makecell{\textbf{Per run}      \\  \textbf{tokens}} 
& \makecell{\textbf{Trajectory} \\  \textbf{tokens}}
& \makecell{\textbf{Completion} \\  \textbf{tokens}}
& \makecell{\textbf{Answer} \\  \textbf{tokens}}

\\
\midrule
Claude Opus 4.5        & $19$ & $27$ & $467{,}000$      & $458{,}000$      & $8{,}500$   & $472$  \\ 
Gemini 3 Flash         & $54$ & $55$ & $5{,}315{,}000$  & $5{,}266{,}000$  & $49{,}000$  & $299$  \\ 
Gemini 3 Pro           & $24$ & $26$ & $667{,}000$      & $652{,}000$      & $16{,}000$  & $191$  \\ 
GPT-5                  & $34$ & $31$ & $1{,}011{,}000$  & $963{,}000$      & $48{,}000$  & $307$  \\ 
GPT-5.2                & $35$ & $45$ & $1{,}042{,}000$  & $1{,}008{,}000$  & $33{,}000$  & $330$  \\ 
GPT-OSS-120B           & $43$ & $26$ & $1{,}635{,}000$  & $1{,}600{,}000$  & $36{,}000$  & $328$  \\ 
Grok 4                 & $31$ & $31$ & $632{,}000$      & $630{,}000$      & $2{,}500$   & $166$  \\ 
Kimi K2 Thinking       & $92$ & $91$ & $1{,}620{,}000$  & $1{,}610{,}000$  & $10{,}000$   & $257$  \\ 

\bottomrule
\end{tabular}
\end{table*}

\subsection{Tools and token use}
The mean number of steps, number of tool calls (note that multiple tools can be used at each step), and token usage for each agent are reported in Table~\ref{tab:trajectory_statistics}. Token usage is split into trajectory tokens (tokens sent from Archipelago, our infra service, to the agent per API call, including the system prompt, conversation history, and any retrieved context) and completion tokens (tokens generated by the agent at each API call, summed across all calls in a trajectory, including response text and any tool call specifications). From the completion tokens, we subset the final answer tokens. The tools most frequently used by agents are given in Appendix~\ref{appendix:tool_usage}.
\vspace{1em}

Agents' use of tokens and tools varies substantially, with Gemini 3 Flash using nearly 5$\times$ as many tokens as GPT-5.2, and about 8$\times$ as many as Gemini 3 Pro. It also uses approximately $54\%$ more steps than GPT-5.2 and $22\%$ more tool calls.  This suggests that although it is effective, Flash is inefficient. At the other end of the leaderboard, Kimi K2 Thinking averages $92$ steps and $91$ tool calls per task and 1.6 million tokens, showing that using more resources does not always result in higher-quality output. Final answer token counts are relatively similar across agents (from $191$ for Gemini 3 Pro to $472$ for Claude Opus 4.5), indicating that agents do not differ substantially in verbosity.
\vspace{1em}

\subsection{Failure analysis}

\paragraph{Tasks with files as output}
For $422$ tasks, agents are required to return a message to console, and for $58$ tasks they are required to create or edit a file (document, spreadsheet, presentation). Agents' scores are lower on tasks that require a file. For both sets of tasks, Gemini 3 Flash performs best, followed by GPT-5.2 and Claude Opus 4.5, with drops of $4.9\%$, $7.2\%$, and $5.0\%$ respectively.
\vspace{1em}

\paragraph{Unwanted file deletion}
Deletions are never requested in the task prompts, and can be seen as undesirable `rogue' behavior by the agent. Claude Opus 4.5, GPT-5 and Kimi K2 Thinking make no deletions. $36$ trajectories contain deletions ($0.12\%$), with GPT-5.2 deleting the most files ($n=21$), followed by Grok 4 ($n=6$), Gemini 3 Flash ($n=5$), and Gemini 3 Pro ($n=2$) and GPT-OSS-120B ($n=2$).
\vspace{1em}

\begin{table}[b]
\small
\caption{Agents' outcomes on APEX\textendash Agents. Each row is $3,840$ trajectories ($1$ x $480$ x $8$) and sums to $100\%$.}
\label{tab:score_breakdown}
\begin{tabular}{
    >{\raggedright\arraybackslash}p{2.4cm} | 
    >{\centering\arraybackslash}p{0.6cm}
    >{\centering\arraybackslash}p{0.9cm}
    >{\centering\arraybackslash}p{0.7cm}
    >{\centering\arraybackslash}p{1.0cm}
}
\toprule
\textbf{Model} & \textbf{Pass} & \textbf{Timeout} & \textbf{Zero} & \textbf{Partial} \\
\midrule
Claude Opus 4.5    & $18.4\%$   & $0.2\%$    & $46.4\%$   & $35.1\%$  \\
Gemini 3 Flash     & $24.0\%$   & $3.2\%$    & $40.3\%$   & $32.6\%$  \\
Gemini 3 Pro       & $18.4\%$   & $0.1\%$    & $46.6\%$   & $34.9\%$  \\
GPT-5              & $18.3\%$   & $1.1\%$    & $46.9\%$   & $33.7\%$  \\
GPT-5.2            & $23.0\%$   & $2.7\%$    & $40.8\%$   & $33.5\%$  \\
GPT-OSS-120B       & $4.7\% $   & $6.5\%$    & $62.2\%$   & $26.6\%$  \\
Grok 4             & $15.2\%$   & $0.0\%$    & $50.5\%$   & $34.3\%$  \\
Kimi K2 Thinking   & $4.0\% $   & $29.8\%$   & $47.1\%$   & $19.1\%$  \\
\bottomrule
\end{tabular}
\end{table}

\paragraph{How agents fail}
On trajectories where agents do not meet all of the criteria, we record how many have their trajectories stopped for exceeding $250$ steps (reflecting a mixture of reasoning limits, planning inefficiencies, and orchestration overhead), how many score $0\%$, and how many achieve partial credit (i.e., score more than $0\%$ and less than $100\%$). See Table~\ref{tab:score_breakdown}. All agents score zero in at least $40\%$ of their runs, showing how difficult the tasks are. This partly reflects our evaluation design, where $92/480$ tasks have only one criterion and $79/480$ have two.
\vspace{1em}

Timeout percentages are substantially higher for the open-source than closed-source models, especially Kimi K2 Thinking, which often doom loops, timing out $29.8\%$ of trajectories. Notably, the top performing agent overall, Gemini 3 Flash, times out the most of the closed-source models ($3.2\%$), followed by the second best agent, GPT-5.2, which times out $2.7\%$ of the time. 
\vspace{1em}

\paragraph{Tool use in passing trajectories}
In trajectories where the agent meets all criteria, code execution accounts for $16.5\%$ of tool calls versus $20.5\%$ in trajectories that do not ($-4.0$ percentage points). Equally, the inspect tool accounts for $3.2\%$ of tool calls versus $6.4\%$ ($-3.3$ percentage points), and listing files accounts for $16.7\%$ versus $12.3\%$ ($+4.4$ percentage points). However, this aggregate analysis does not control for differences in task difficulty, nor for differences in agents’ problem-solving approaches. Instead, to make a head-to-head comparison, we use the $937$ of $3{,}840$ agent–task pairs (reflecting $7{,}496$ trajectories) where the agent has at least one run that passes the task and one run that fails. 
\vspace{1em}

In a head-to-head comparison, trajectories that pass all criteria use slightly more unique tools (an increase of $0.22$ per trajectory) and are $3$ percentage points more likely to use code at least once. They also have substantially fewer steps ($-5.95$) and make far fewer tool calls overall ($-5.66$). Successful trajectories likely avoid repeatedly using unproductive tools (i.e., ``doom looping'').
\vspace{1em}

\section{Conclusion}
The APEX\textendash Agents benchmark evaluates whether frontier agents are capable of performing long-horizon, cross-application tasks in professional services. We find that agents have substantial headroom to improve, with the top-performing agents (Gemini 3 Flash, GPT-5.2, Claude Opus 4.5) all scoring under $25\%$ when measured with Pass@1, and no more than $40\%$ when measured with both Pass@8 and mean score. Failures vary -- from running out of steps, to failing to meet any criteria, or achieving partial credit (which indicates some useful output). We also see variance across runs, and a substantial drop from Pass@8 to Pass\textasciicircum{8}, demonstrating that agents are capable of executing complex professional services work, but often do so inconsistently.
\vspace{1em}

To enable open research, the dataset for APEX\textendash Agents is available open-source, along with Archipelago, our infra service. In future benchmarks, we plan to expand the horizon of tasks, the complexity and depth of worlds, and the value that the benchmark represents.
\vspace{1em}

\section{Acknowledgments}
We thank all the experts from the Mercor marketplace who contributed to APEX\textendash Agents and team members at Mercor who gave feedback on the design of APEX\textendash Agents and supported the project. We thank Jonathan Frankle for his advice.

\bibliography{anthology,custom}

@inproceedings{styles2024workbench,
title={WorkBench: a Benchmark Dataset for Agents in a Realistic Workplace Setting},
author={Olly Styles and Sam Miller and Patricio Cerda-Mardini and Tanaya Guha and Victor Sanchez and Bertie Vidgen},
booktitle={First Conference on Language Modeling},
year={2024},
url={https://openreview.net/forum?id=4HNAwZFDcH}
}

@misc{meimandi2025measurementimbalanceagenticai,
      title={{The Measurement Imbalance in Agentic AI Evaluation Undermines Industry Productivity Claims}}, 
      author={Kiana Jafari Meimandi and Gabriela Aránguiz-Dias and Grace Ra Kim and Lana Saadeddin and Allie Griffith and Mykel J. Kochenderfer},
      year={2025},
      eprint={2506.02064},
      archivePrefix={arXiv},
      primaryClass={cs.CY},
      url={https://arxiv.org/abs/2506.02064}, 
}

@misc{froger2025arescalingagentenvironments,
      title={{ARE: Scaling Up Agent Environments and Evaluations}}, 
      author={Romain Froger and Pierre Andrews and Matteo Bettini and Amar Budhiraja and Ricardo Silveira Cabral and Virginie Do and Emilien Garreau and Jean-Baptiste Gaya and Hugo Laurençon and Maxime Lecanu and Kunal Malkan and Dheeraj Mekala and Pierre Ménard and Gerard Moreno-Torres Bertran and Ulyana Piterbarg and Mikhail Plekhanov and Mathieu Rita and Andrey Rusakov and Vladislav Vorotilov and Mengjue Wang and Ian Yu and Amine Benhalloum and Grégoire Mialon and Thomas Scialom},
      year={2025},
      eprint={2509.17158},
      archivePrefix={arXiv},
      primaryClass={cs.AI},
      url={https://arxiv.org/abs/2509.17158}, 
}

@misc{kapoor2024aiagentsmatter,
      title={{AI Agents That Matter}}, 
      author={Sayash Kapoor and Benedikt Stroebl and Zachary S. Siegel and Nitya Nadgir and Arvind Narayanan},
      year={2024},
      eprint={2407.01502},
      archivePrefix={arXiv},
      primaryClass={cs.LG},
      url={https://arxiv.org/abs/2407.01502}, 
}

@misc{dechezelles2025browsergymecosystemwebagent,
      title={{The BrowserGym Ecosystem for Web Agent Research}}, 
      author={Thibault Le Sellier De Chezelles and Maxime Gasse and Alexandre Drouin and Massimo Caccia and Léo Boisvert and Megh Thakkar and Tom Marty and Rim Assouel and Sahar Omidi Shayegan and Lawrence Keunho Jang and Xing Han Lù and Ori Yoran and Dehan Kong and Frank F. Xu and Siva Reddy and Quentin Cappart and Graham Neubig and Ruslan Salakhutdinov and Nicolas Chapados and Alexandre Lacoste},
      year={2025},
      eprint={2412.05467},
      archivePrefix={arXiv},
      primaryClass={cs.LG},
      url={https://arxiv.org/abs/2412.05467}, 
}

@misc{vidgen2025aiproductivityindexapex,
      title={{The AI Productivity Index (APEX)}}, 
      author={Bertie Vidgen and Abby Fennelly and Evan Pinnix and Julien Benchek and Daniyal Khan and Zach Richards and Austin Bridges and Calix Huang and Kanishka Sahu and Abhishek Kottamasu and Bo Ma and Ben Hunsberger and Isaac Robinson and Akul Datta and Chirag Mahapatra and Dominic Barton and Cass R. Sunstein and Eric Topol and Brendan Foody and Osvald Nitski},
      year={2025},
      eprint={2509.25721},
      archivePrefix={arXiv},
      primaryClass={econ.GN},
      url={https://arxiv.org/abs/2509.25721}, 
}
\bibliographystyle{acl_natbib}

\appendix
\section{Archipelago}\label{sec:archipelago}
Archipelago is open-source infrastructure for running and evaluating AI agents against RL environments. It has three components: (1) the Environment, a containerized sandbox that exposes multiple applications through a unified Model Context Protocol gateway; (2) an Agents runner that runs LMs with different agentic harnesses (for this benchmark, a ReAct toolbelt); and (3) a Grading system that evaluates agent outputs by comparing before-and-after snapshots of the world, using verifiers. All three components are Docker containers that can run on any orchestration platform (e.g., Kubernetes, Modal).
\vspace{1em}

\begin{table}[t]
\centering
\small
\caption{Count of how many times each workflow appears in the tasks.}
\label{tab:workflow_distribution}

\begin{tabular}{l|lc}
\toprule
\textbf{Job} & \textbf{Workflow} & \textbf{Count} \\
\midrule

\multirow{9}{*}{\makecell[l]{Investment\\banking\\analyst}}
& Comparables & 16 \\
& DCF & 42 \\
& Debt Model & 6 \\
& LBO & 12 \\
& Market / Sector Research & 3 \\
& Merger Model & 7 \\
& \makecell[l]{Sensitivity\\Analysis} & 46 \\
& Valuation Analysis & 28 \\
\midrule

\multirow{7}{*}{\makecell[l]{Management\\consultant}}
& \makecell[l]{Benchmarking / Competitive\\Analysis} & 26 \\
& Cost Benefit Analysis & 11 \\
& Market Sizing, TAM, SAM & 14 \\
& Operations Analysis & 23 \\
& \makecell[l]{Scenario/Sensitivity\\Analysis} & 35 \\
& Strategy Recommendations & 5 \\
& \makecell[l]{Survey / Interview\\Analysis} & 31 \\
& \makecell[l]{Variance / Performance\\Analysis} & 15 \\
\midrule

\multirow{9}{*}{\makecell[l]{Corporate\\lawyer}}
& Compliance Program Review & 16 \\
& Contract Review & 30 \\
& Due Diligence & 18 \\
& Internal Investigations & 3 \\
& Legal Research & 47 \\
& Litigation Strategy & 8 \\
& Motion Drafting & 6 \\
& Risk Assessment & 24 \\
& Other & 8 \\
\bottomrule
\end{tabular}
\end{table}

\section{Time spent at work by professionals} \label{appendix:expert_time_allocation}
The time allocation reported by participants in the APEX Survey is given in Table~\ref{tab:time_spent}.

\begin{table*}[tb]
\small
\centering
\caption{Time spent by professionals at work, based on responses from participants in the APEX Survey ($n=227$).}
\label{tab:time_spent}
\begin{tabular}{lcccc}
\toprule
 & \textbf{All jobs}
 & \textbf{Investment banking analyst}
 & \textbf{Management consultant} 
  & \textbf{Lawyer} \\
\midrule
Core activities        & 47.4\%  & 47.2\%  & 43.5\%  & 50.0\% \\
Learning               & 9.2\%   & 8.3\%   & 7.1\%   & 11.4\% \\
Admin                  & 10.5\%  & 10.4\%  & 10.1\%  & 11.1\% \\
Comms                  & 13.5\%  & 13.5\%  & 14.7\%  & 12.8\% \\
Meetings               & 14.3\%  & 14.7\%  & 18.7\%  & 10.2\% \\
Non-productive         & 5.1\%   & 6.0\%   & 5.9\%   & 4.5\%  \\
\midrule
Total                  & 100\%  & 100\%  & 100\%  & 100\%  \\
\bottomrule
\end{tabular}
\end{table*}

\begin{table}[t]
\small
\centering
\caption{Breakdown of time spent on core activities by professionals at work, based on responses from participants in the APEX Survey ($n=227$). We show the five most reported categories of work for each job.}
\label{tab:time_spent_detailed}
\begin{tabular}{p{2.2cm} p{4.5cm}}
\toprule
\textbf{Job} & \textbf{Top activities (share of time)} \\
\midrule
\multirow{5}{*}{\makecell[l]{Investment\\banking\\analyst}} &
Financial modelling (23.5\%)\\
& Research (13.6\%)\\
& Quantitative data analysis (12.8\%) \\
& Prep \& give presentations (11.9\%) \\
& Managing stakeholders (6.0\%) \\
\midrule
\multirow{5}{*}{\makecell[l]{Management\\consultant}} &
Quantitative data analysis (17.8\%) \\
& Prep \& give presentations (15.1\%) \\
& Customer contact (10.4\%) \\
& Research (9.9\%)\\
& Staff feedback \& review (8.3\%) \\
\midrule
\multirow{5}{*}{Lawyer} &
Research (24.9\%)\\
& Docs, reports, memos (16.4\%)\\
& Contracts and legal docs (14.7\%)\\
& Customer contact (10.9\%)\\
& Other legal work (9.4\%) \\
\bottomrule
\end{tabular}
\end{table}

\section{Distribution of workflows}
The distribution of workflows over the tasks in APEX\textendash Agents is given in Table~\ref{tab:workflow_distribution}. The five most reported categories of work for each job are given in Table~\ref{tab:time_spent_detailed}.

\section{Agent setup} \label{appendix:agent_setup}

\subsection{ReAct toolbelt}
We use a looped toolbelt, following the ReAct paradigm: Reasoning and Acting are interleaved in a single loop. At each step, the agent observes the current state, reasons about the next action, executes tool(s), and repeats. 
\vspace{1em}

The agent requires an explicit final\_answer tool call to complete. This ensures intentional termination with structured output (i.e., an answer + status). To ensure tools do not bloat model context, we implement a toolbelt approach where the agent initially only has access to (1) meta-tools for discovering, describing, and managing available tools and (2) task planning (i.e., creating a to-do list with batch support). The agent also includes context summarization that is triggered when $70\%$ of the context window is utilized and retains the last $10$ messages. This efficient context window management minimizes the risk that agents fail a task because they run out of tokens. See our GitHub repository.
\vspace{1em}

\subsection{Agent system prompt}
All agents use the same system prompt:
\vspace{1em}

\begin{quote}
You are an agent that completes tasks independently.
Use the tools provided to you to complete the task to the best of your ability.
You should use the \texttt{code\_exec} tool when needed, such as when calculating values.
When calculating numbers, unless specified otherwise, use the exact values without rounding them.

You must attempt to execute the task. You cannot ask for help or further clarification.

For every tool except the \texttt{code\_exec} tool, you may assume that all relevant files are located under the root path \texttt{/}. For the \texttt{code\_exec} tool, however, you must explicitly use \texttt{/filesystem/} as the root path to locate all relevant files.
\end{quote}

\section{Model details} \label{appendix:model_configs}
An overview of the models tested against APEX\textendash Agents is given in Table~\ref{tab:model_configs}.

\begin{table*}[tb]
\small
\centering
\caption{Configs of the models tested against the APEX\textendash Agents benchmark.}
\label{tab:model_configs}
\begin{tabular}{l|llll}
\toprule
\textbf{Model name} &
\textbf{Provider} &
\textbf{Access} &
\textbf{Thinking settings} &
\textbf{Other configs} \\
\midrule
Claude Opus 4.5    & Anthropic         & Closed-source   & High                 & \makecell[l]{Temperature = 1.0 \\ Max tokens = $64,000$}  \\
\midrule
Gemini 3 Flash     & Google DeepMind   & Closed-source   & High                 & \makecell[l]{Temperature = 1.0 \\ Verbosity = Medium \\ Max tokens = $65,536$} \\
\midrule
Gemini 3 Pro       & Google DeepMind   & Closed-source   & High                 & \makecell[l]{Temperature = 1.0 \\ Verbosity = Medium \\ Max tokens = $65,536$} \\
\midrule
GPT-5              & OpenAI            & Closed-source   & High                 & \makecell[l]{Verbosity = Medium \\ Max tokens = $128,000$} \\
\midrule
GPT-5.2            & OpenAI            & Closed-source   & High                 & \makecell[l]{Verbosity = Medium \\ Max tokens = $128,000$}  \\
\midrule
GPT-OSS-120B       & OpenAI            & Open-source     & High                 & \makecell[l]{Temperature = $0.7$ \\ Max tokens = $131,072$}  \\
\midrule
Grok 4             & xAI               & Closed-source   & [On by default]      & \makecell[l]{Temperature = $0.8$ \\ Max tokens = $16,000$} \\
\midrule
Kimi K2 Thinking   & Moonshot AI       & Open-source     & [On by default]      & \makecell[l]{Temperature = $0.7$ \\ Max tokens = $262,144$} \\
\bottomrule
\end{tabular}
\end{table*}

\section{Significance tests on Pass@1 scores} \label{sec:sig_tests}
Pairwise scores for models tested against APEX\textendash Agents, using McNemar's exact test with a Benjamini–Hochberg correction, are given in Table~\ref{tab:sig_tests}.

\begin{table*}
\small
\centering
\caption{Pairwise Pass@1 scores with Benjamini--Hochberg correction (m=28).}
\label{tab:sig_tests}
\begin{tabular}{ll|cc}
\toprule
\textbf{Model A} &\textbf{Model B} & \textbf{p (McNemar)} & \textbf{q (BH)} \\
\midrule
Kimi-K2-Thinking & gemini-3-flash-preview & 5.68e-23 & 1.59e-21 \\
gemini-3-flash-preview & gpt-oss-120b & 6.51e-21 & 9.11e-20 \\
Kimi-K2-Thinking & gpt-5.2 & 3.21e-20 & 2.99e-19 \\
gpt-5.2 & gpt-oss-120b & 8.67e-19 & 6.07e-18 \\
Kimi-K2-Thinking & claude-opus-4-5 & 3.60e-13 & 2.02e-12 \\
claude-opus-4-5 & gpt-oss-120b & 1.24e-12 & 5.79e-12 \\
gemini-3-pro-preview & gpt-oss-120b & 1.85e-12 & 7.39e-12 \\
Kimi-K2-Thinking & gemini-3-pro-preview & 4.03e-12 & 1.41e-11 \\
gpt-5 & gpt-oss-120b & 7.24e-10 & 2.04e-09 \\
Kimi-K2-Thinking & gpt-5 & 7.29e-10 & 2.04e-09 \\
gemini-3-flash-preview & grok-4 & 2.92e-08 & 7.43e-08 \\
Kimi-K2-Thinking & grok-4 & 3.64e-08 & 8.07e-08 \\
gpt-oss-120b & grok-4 & 3.75e-08 & 8.07e-08 \\
gpt-5.2 & grok-4 & 3.69e-06 & 7.38e-06 \\
gemini-3-flash-preview & gpt-5 & 3.30e-05 & 6.15e-05 \\
gpt-5 & gpt-5.2 & 0.0002 & 0.0004 \\
gemini-3-flash-preview & gemini-3-pro-preview & 0.0003 & 0.0004 \\
claude-opus-4-5 & gemini-3-flash-preview & 0.0005 & 0.0008 \\
gemini-3-pro-preview & gpt-5.2 & 0.0065 & 0.0096 \\
claude-opus-4-5 & gpt-5.2 & 0.0073 & 0.0102 \\
claude-opus-4-5 & grok-4 & 0.0503 & 0.0670 \\
gemini-3-pro-preview & grok-4 & 0.0893 & 0.1140 \\
claude-opus-4-5 & gpt-5 & 0.3480 & 0.4240 \\
gpt-5 & grok-4 & 0.3910 & 0.4560 \\
gemini-3-pro-preview & gpt-5 & 0.4970 & 0.5560 \\
gemini-3-flash-preview & gpt-5.2 & 0.5550 & 0.5970 \\
claude-opus-4-5 & gemini-3-pro-preview & 0.9180 & 0.9520 \\
Kimi-K2-Thinking & gpt-oss-120b & 1.0000 & 1.0000 \\
\bottomrule
\end{tabular}
\end{table*}

\section{Tools most frequently used by agents}\label{appendix:tool_usage}
Rounded to the nearest thousand, the ten tools most frequently used by agents are: Code execution ($256,000$), Add tool to the toolbelt ($200,000$), List files in the file system ($164,000$), Read spreadsheet tab ($127,000$), Search the PDF ($86,000$), Inspect tool with the toolbelt ($78,000$), Read PDF ($55,000$), Read document content ($45,000$), List tabs in a spreadsheet ($42,000$), and Read an image ($37,000$).

\end{document}